\documentclass[10pt,twocolumn,letterpaper]{article}

\usepackage{cvpr}
\usepackage{times}
\usepackage{epsfig}
\usepackage{graphicx}
\usepackage{amsmath}
\usepackage{amssymb}

\usepackage{color}
\definecolor{ChrisCol}{RGB}{255,50,50}
\definecolor{MichaelCol}{RGB}{100,120,240}

\usepackage{amsmath}
\usepackage{amsfonts}
\usepackage{amssymb}
\usepackage{algorithm}
\usepackage[noend]{algpseudocode}

\newcommand{\reals}{\mathbb{R}}
\newcommand{\bv}[1]{\mathbf{#1}}

\usepackage{lineno}
\usepackage{multirow}
\usepackage{array}
\usepackage{stackrel}
\usepackage{tabularx}
\usepackage{subfig}
\usepackage{url}
\newcommand{\argmin}[1]{\underset{#1}{\operatorname{arg}\,\operatorname{min}}\;}

\definecolor{darkgreen}{rgb}{0.25,0.65,0.3}
\definecolor{darkred}{rgb}{0.85,0.25,0.3}
\usepackage[pagebackref=true,breaklinks=true,letterpaper=true,colorlinks,bookmarks=false,citecolor=darkgreen,linkcolor=darkred]{hyperref}

\cvprfinalcopy 

\def\cvprPaperID{751} 

\ifcvprfinal\fi
\begin{document}

\title{Combining Markov Random Fields and Convolutional Neural Networks for Image Synthesis}

\author{Chuan Li\\
Johannes Gutenberg University\\
Mainz, Germany\\
{\tt\small chuli@uni-mainz.de}
\and
Michael Wand\\
Johannes Gutenberg University\\
Mainz, Germany\\
{\tt\small wandm@uni-mainz.de}
}

\maketitle

\begin{abstract}
This paper studies a combination of generative Markov random field (MRF) models and discriminatively trained deep convolutional neural networks (dCNNs) for synthesizing 2D images. The generative MRF acts on higher-levels of a dCNN feature pyramid, controling the image layout at an abstract level.
We apply the method to both photographic and non-photo-realistic (artwork) synthesis tasks. The MRF regularizer prevents over-excitation artifacts and reduces implausible feature mixtures common to previous dCNN inversion approaches, permitting synthezing photographic content with increased visual plausibility. Unlike standard MRF-based texture synthesis, the combined system can both match and adapt local features with considerable variability, yielding results far out of reach of classic generative MRF methods.
%
%
\end{abstract}

\section{Introduction}

The problem of synthesizing content by example is a classic problem in computer vision and graphics. It is of fundamental importance to many applications, including creative tools such as high-level interactive photo editing \cite{Agarwala04,Barnes09,Hertzmann01}, as well as scientific applications, such as generating stimuli in psycho-physical experiments~\cite{Gatys2015b}.

\begin{figure}
	\centering
		\includegraphics[width=1\linewidth]{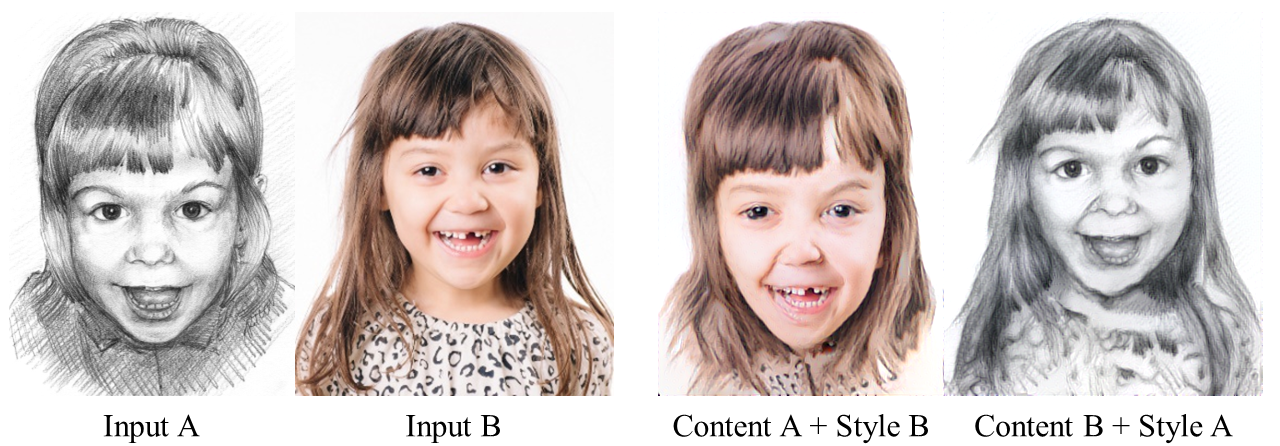}
		\caption{By combining deep convolutional neural network with MRF prior, our method can transfer both photorealistic and non-photorealistic styles to new images. Images credited to flickr users \emph{mricon} (A) and \emph{Peter Dahlgren} (B).}
	\label{fig:teaser}
\end{figure}

In this paper, we specifically consider the problem of data-driven image synthesis: Given an example image, we want to fully automatically create a variant of the example image that looks similar but differs in structure. The intended deviation is controlled by additional constraints provided by the user, ranging from just changing image dimensions to detailed layout specifications. Concretely, we implement this by splitting up the input into a \emph{``style''} image and a \emph{``content''} image~\cite{Gatys2015b,Hertzmann01}. The first describes the building blocks the image should be made of, the second constrains their layout. Figure~\ref{fig:teaser} shows an example of style transferred images, where the input images are shown on the left. Our results are are shown on the right. Notice our method produces plausible results for both art to photo and photo to art transfers. In particular, see how meso-structures in the style images, such as the mouth and eyes, are intentionally reused in the synthesized images.

The classic data-driven approach to generative image modeling is based on Markov random fields (MRFs): We assume that the most relevant statistical dependencies in an image are present at a local level and learn a distribution over the likelihood of local image patches by considering all local $k \times k$ pixel patches in the example image(s). Usually, this is done using a simple nearest-neighbor scheme \cite{Efros99}, and inference is performed by approximate MRF inference \cite{Efros01,Kwatra05,Kwatra03} or greedy approximations \cite{Barnes09,Efros99,Hertzmann01,Wei00}.

\begin{figure*}[t]
	\centering
		\includegraphics[width=1\linewidth]{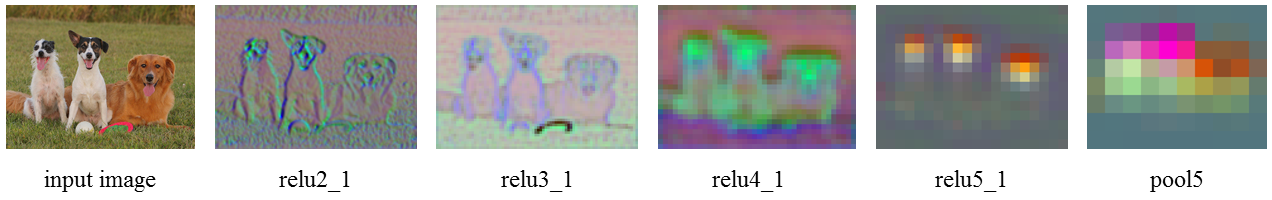}
		\caption{The input image is encoded by the VGG network (pixel colors show a 3D PCA embedding of the high-dimensiontal feature space). Related image content is mapped to semi-distributed, approximately spatially coherent feature constellations of increasing invariance \protect\cite{Zeiler14}. Input image credited to flickr user \emph{Emery Way}.}
	\label{fig:PCAface}
\end{figure*}

A critical limitation of MRFs texture synthesis is the difficulty of learning the distribution of plausible image patches from example data. Even the space of local $k \times k$ image patches (typically: $k \approx 5...31$) is already way too high-dimensional to be covered with simple sampling and nearest-neighbor estimation. The results are mismatched local pieces, which are subsequently stitched~\cite{Kwatra03} or blended~\cite{Kwatra05} together in order to minimize the perceptual impact of the lack of ability to generalize. The missing ingredient is a strong scheme for interpolating and adapting images from very sparse example sets of sample patches. 

In terms of invariance and ability to generalize, discriminatively trained deep convolutional neural networks have recently made dramatic impact~\cite{Krizhevsky2012,Simonyan14c}. They are able to recognize complex classes of image features, modeling non-linear deformations and appearance variations far beyond the abilities of simple nearest neighbors search.

However, the discriminative design poses a problem: The corresponding dCNNs compress image information progressively over multiple pooling layers to a very coarse representation, which encodes semantically relevant feature descriptors in a semi-distributed (not spatially localized) fashion (Figure~\ref{fig:PCAface}). While an inverse process can be defined \cite{Denton15,Dong15,Gatys15,Gatys2015b,Gauthier15,Mahendran15,Xu15,Zeiler14}, it reamins difficult to control: For example, simply maximizing the class-excitation of the network leads to hallucinatory patterns~\cite{Mordvintsev15}. Rather than that, we need to reproduce the correct statistics for neural encoding in the synthesis images. 


Addressing this problem, Gatys et al.~\cite{Gatys15,Gatys2015b} have recently demonstrated remarkable results for transferring styles to guiding ``content'' images: Their method uses the filter pyramid of the VGG network~\cite{Simonyan14c} as a higher-level representation of images, benefitting from the vast knowledge acquired through training the dCNN on millions of photographs. Then, feature layout is simply controlled by penalizing the difference of the high-level neural encoding of the newly synthesized image to that of the content image. Further, the process is regularized by matching feature statistics of the ``style'' image and the newly synthesized picture by matching the correlations across the various filter channels, captured by a Gram matrix. The method yields very impressive results for applying artistic styles of paintings to photographs~\cite{Gatys15}. However, strict local plausibility remains difficult. In particular, using photographs as styles does not yield plausible results because only \emph{per-pixel} feature correlations are captured at different layers and the spatial layout is constrained too weakly.

Our paper augments their framework by replacing the bag-of-feature-like statistics of Gram-matrix-matching by an MRF regularizer that maintains local patterns of the ``style'' exemplar: MRFs and dCNNs are a canonical combination --- both models crucially rely on the assumption of locally correlated information and translational invariance. This equips the encoding of features in a dCNN with approximate Markovian consistency properties: Local patches have characteristic arrangements of feature activations to describe objects, and higher-up encoding becomes more invariant under in-class variation (Figure~\ref{fig:PCAface}).
We therefore use the generative MRF model on such higher levels of the network (\textit{relu3\_1} and \textit{relu4\_1} of the 19-layer VGG network\footnote{Notice that in Figure~\ref{fig:PCAface} layer \textit{relu5\_1} shows the most discriminative encoding for single pixels. However, in practice we found using $3 \times 3$ patches at layer \textit{relu4\_1} produce the best synthesis results. Intuitively, using patches from a slightly lower layer has similar matching performance, but permits overlapped MRFs and increased details in synthesis.}). This prescribes a plausible local layout of objects in terms of more abstract categories and, importantly, tries to ensure a consistent encoding of these higher-level features. The actual task of generalizing within object categories and plausible blending is then performed by the dCNN's lower levels via inversion~\cite{Mahendran15}.

Technically, we implement the additional MRF prior by an additional energy term that models Markovian consistency of the upper layers of the dCNN feature pyramid. We then utilize the EM algorithm of Kwatra et al.~\cite{Kwatra05} for MRF optimization: It easily integrates in the variational framework. Further, we will show that higher-level neural encodings are more perceptually linear, which matches well with the linear blending approach of the M-step.

We apply our method to a number of photo-realistic and non-photo-realistic image synthesis tasks, and show it is able to generalize among image patches far beyond the abilities of classic MRFs. In style-transfer scenarios, the combined method additionally benefits from the abilities of the dCNN to match semantically related image portions automatically, without user annotations. In comparison to previous methods that invert dCNNs, the MRF prior improves local plausibility of the feature layouts, avoiding hallucinatory artifacts and usually providing a more plausible meso-structure than the statistical approach of Gatys et al.~\cite{Gatys15}. In particular, we can strongly improve the plausibility of synthesizing photographs, which was not possible with the previous methods.





\section{Related Work}


\textbf{Image synthesis with neural networks:} 
The success of dCNNs in discriminative tasks~\cite{ILSVRC15} has also raised interest in generative variants. Zeiler et al.~\cite{Zeiler14} introduce a deconvolutional network to back-project neuron activations to pixels. Similarly, Mahendran and  Vedaldi~\cite{Mahendran15} reconstruct images from the neural encoding in intermediate layers. The work of Gatys et al~\cite{Gatys15}, detailed above, can also be employed in unguided settings~\cite{Gatys2015b}, outperforming traditional parametric texture synthesis which only uses a linear feature bank and no statistical priors~\cite{Portilla00}. 

A further approach is the framework of generative adversarial networks~\cite{Goodfellow2014}. Here, two networks, one as the discriminator and other as the generator iteratively improve each other by playing a minnimax game. In the end the generator is able to produces more natural images than naive image synthesis. However, in many cases the output quality is still rather limited. Gauthier et al.~\cite{Gauthier15} extend this model by a Laplacian pyramid. This leads to clear improvement on output quality. Nonetheless, training for large images remains expensive and the results often still lack structure. Denton et al.~\cite{Denton15} extend the model to a conditional setting, limited to generating faces. It is also possible to re-train networks for specific generative tasks, such as image deblur~\cite{Xu15}, super-resolution ~\cite{Dong15}, and class visualization~\cite{Lu15}. 

\textbf{MRF-based image synthesis:}
MRFs are the classic framework for for non-parametric image synthesis~\cite{Efros99}.
As explained in the introduction, a key issue is adapting local patches beyond simple stitching~\cite{Kwatra03} or blending~\cite{Kwatra05}, and our paper focuses on this issue. Aside from this, MRF models suffer from a second, significant limitation: Local image statistics is usually not sufficient for capturing complex image layouts at a global scale. While local details appear plausible, the global arrangement often merely resembles an unstructured ``texture soup''. Multi-resolution synthesis~\cite{Hertzmann01,Kwatra05,Wei00} provides some improvement here (and we adapt this in our method, too), but a principled solution requires additional high-level constraints. These can be either explicitly provided by the user~\cite{Agarwala04,Barnes09,Efros01,Gatys15,Hertzmann01,Lee10,Xie07}, or learned from non-local image statistics~\cite{He12,Li15,Zhang13}. Long range correlations have also been modeled by spatial LTSM neural networks; results so far are still limited to semi-regular textures~\cite{Theis2015}.
 Our paper opts for the first, simpler solution of explicit layout constraints through a ``content'' image~\cite{Gatys15,Hertzmann01} --- in principle, learning of global layout constraints is mostly orthogonal to our approach.

\section{Model}
We now discuss our combined MRFs and dCNNs model for image synthesis. We assume that we are given a style image, denoted by $\bv{x}_s \in \reals^{w_s \times h_s}$, and a content image $\bv{x}_c \in \reals^{w_c \times h_c}$ for guidance. The (yet unknown) synthesized image is denoted by $\bv{x} \in \reals^{w_c \times h_c}$. We transfer the style of $\bv{x}_s$ into the layout of $\bv{x}_c$ by making the high-level neural encoding of $\bv{x}$ similar to $\bv{x}_c$, but using local patches similar to those of $\bv{x}_s$. The latter is the MRF prior that maintains the encoding of the style.
%
%
Formally, $\bv{x}$ minimizes the following energy function:
%
\begin{eqnarray}
\bv{x}&= &\argmin{x}E_{s}(\Phi(\bv{x}), \Phi(\bv{x}_{s})) + \nonumber \\&& \alpha_{1}E_{c}(\Phi(\bv{x}), \Phi(\bv{x}_{c})) + \alpha_{2}\Upsilon(\bv{x})
\label{eq:energy}
\end{eqnarray}
%
$E_{s}$ denotes the style loss function (MRFs constraint), where $\Phi(\bv{x})$ is $\bv{x}$'s feature map from some layer in the network. 
%
$E_{c}$ is the content loss function. It computes the squared distance between the feature map of the synthesis image and that of the content guidance image $\bv{x}_{c}$. As shown in~\cite{Gatys15,Mahendran15}, minimizing $E_{c}$ generates an image that is contextually related to $\bv{x}_{c}$. The additional regularizer $\Upsilon(\bv{x})$ is a smoothness prior on the reconstruction. Next, we explain how to define these terms in details.
%
%
%
%

\textbf{MRFs loss function}: Let $\Psi(\Phi(\bv{x}))$ denote the list of all local patches extracted from $\Phi(\bv{x})$ -- a specified set of feature maps of $\bv{x}$. Each ``neural patch'' is indexed as $\Psi_i( \Phi(\bv{x}) )$ and of the size $k \times k \times C$, where $k$ is the width and height of the patch, and $C$ is the number of channels for the layer where the patch is extracted from.
We set the energy function to
\begin{equation}
E_{s}(\Phi(\bv{x}), \Phi(\bv{x}_{s})) = \sum_{i = 1}^{m}||\Psi_i(\Phi(\bv{x})) - \Psi_{NN(i)}(\Phi(\bv{x}_{s}))||^{2}
\label{eq:mrf}
\end{equation}
Here $m$ is the cardinality of $\Psi(\Phi(\bv{x}))$.
For each patch $\Psi_i(\Phi(\bv{x}))$ we find its best matching patch $\Psi_{NN(i)}(\Phi(\bv{x}_{s}))$ using normalized cross-correlation over all $m_s$ example patches in $\Psi(\Phi(\bv{x}_{s}))$: 
\begin{equation}
NN(i) := \argmin{j=1,...,m_s}\frac{\Psi_i(\Phi(\bv{x}))\cdot\Psi_j(\Phi(\bv{x}_{s}))}{|\Psi_i(\Phi(\bv{x}))|\cdot|\Psi_j(\Phi(\bv{x}_{s}))|}
\label{eq:ncorr}
\end{equation}
We use normalized cross-correlation to achieves stronger invariance. The matching process can be efficiently executed by an additional convolutional layer (explained in the implement details). Notice although we use normalized cross-correlation to find the best match, their Euclidean distance is minimized in Equation~\ref{eq:mrf} for producing an image that is visually close to the reference style.  

\textbf{Content loss function}: $E_{c}$ guides the content of the synthesized image by minimizing the squared Euclidean distance between $\Phi(\bv{x})$ and $\Phi(\bv{x}_{c})$:
\begin{equation}
E_{c}(\Phi(\bv{x}), \Phi(\bv{x}_{c})) = ||\Phi(\bv{x}) - \Phi(\bv{x}_{c})||^{2}
\label{eq:content}
\end{equation}

\textbf{Regularizer}: There is significant amount of low-level image information discarded during the discriminative training of the network. In consequence, reconstructing an image from its neural encoding can be noisy and unnatural. For this reason, we penalize the squared gradient norm~\cite{Mahendran15} to encourage smoothness in the synthesized image:
\begin{equation}
\Upsilon(\bv{x}) = \sum_{i, j}\left((x_{i, j + 1} - x_{i, j})^{2} + (x_{i + 1, j} - x_{i, j})^{2}\right)
\label{eq:tv}
\end{equation}
\textbf{Minimization}: We minimize Equation~\ref{eq:energy} using back-propagation with Limited-memory BFGS. In particular, the gradient of $E_{s}$ with respect to the feature maps is the element-wise difference between $\Phi(\bv{x})$ and their MRFs based reconstruction using patches from $\Phi(\bv{x}_{s})$. Such a reconstruction is essentially a texture optimization process~\cite{Kwatra05} that uses neural patches instead of pixel patches. It is crucially important to optimize this MRF energy at the neural level, as the traditional pixel based texture optimization will not be able produce results of comparable quality.

\begin{figure}[t]
	\centering
	\includegraphics[width=0.75\linewidth]{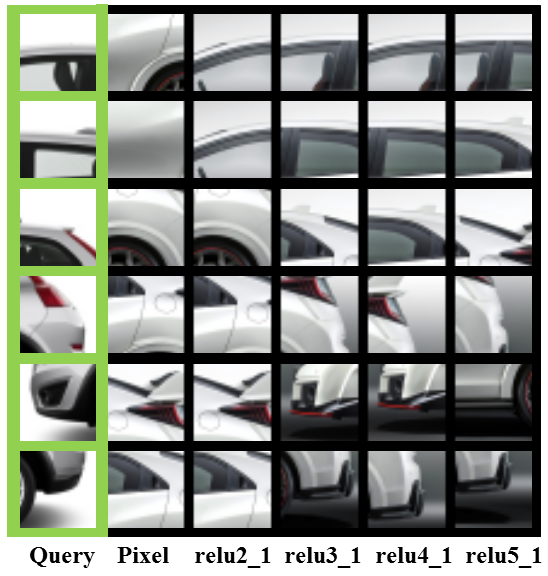}
	\caption{Comparison of patch matching at different layers of a VGG network.}\label{fig:analysis_matching}
\end{figure}

\textbf{Weight} The $\alpha_{1}$ and $\alpha_{2}$ are weights for the content constraint and the natural image prior, respectively. We set $\alpha_{1} = 0$ for non-guided synthesis. By default we set $\alpha_{1} = 1$ for style transfer, while user can fine tune this value to interpolate between the content and the style. $\alpha_{2}$ is fixed to $0.001$ for all cases.

\section{Analysis}
Our key insight is that combining MRF priors with dCNN can significantly improve synthesis quality. This section provides a detailed analysis of our method from three perspectives: we first show compared to pixel values, neural activation leads to better patch matching and blending. We then show how MRFs can further improve the results.

\subsection{Neural Matching}
A key component of non-parametric image synthesis is to match the synthesized data with the example. (Figure~\ref{fig:analysis_matching} is a toy example that shows neural activation gives better matching than pixels. The task is to match two different car images. The first column contains the query patches from one car; every other column shows the best matching in the other car, found at different feature maps (including the pixel layer).

It is clear that patch matching using pixels or neural activation at the lower layers (such as \textit{relu2\_1}) is sensitive to appearance variation. The neural activations at layers \textit{relu3\_1} and \textit{relu4\_1} give better results. The top layers (\textit{relu5\_1}) seem to decrease the match quality due to the increasing invariance against appearance variation. This is not surprising because the features at middle layers are usually trained for recognizing object parts, as discussed in~\cite{Zeiler14}. For these reason, we use neural patches at layers \textit{relu3\_1} and \textit{relu4\_1} as MRFs to constrain the synthesis process.

\begin{figure}[t]
	\centering
	\includegraphics[width=0.95\linewidth]{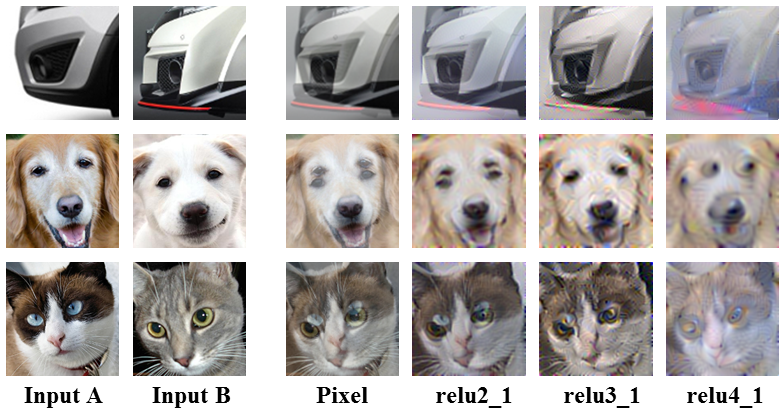}
	\caption{Linear blending behave differently in pixel space and in neural space. Input photos credited to: \emph{rockysretreat.com} (dog A), \emph{wall321.com} (dog B), flickr user \emph{Jsome1} (cat A), flickr user \emph{MendocinoAnimalCare} (cat B).}\label{fig:analysis_blending}
\end{figure}

\subsection{Neural Blending}
The least-squared optimization for minimizing the texture term ($E_{s}$ in Equation~\ref{eq:mrf}) leads to a linear blending operation for overlapping patches. Here we show that blending neural patches often works better than directly blending pixel patches. Specifically, we compare two groups of blending results: The first method is to directly blend the pixels of two input patches. The second method passes these patches through the network and blend their neural activations at different layers. For each layer, we then reconstruct the blending result back into the pixel space using the method described in~\cite{Mahendran15}.

Figure~\ref{fig:analysis_blending} compares the results of these two methods. The first two columns are the input patches A and B for blending. They are intentionally chosen to be semantically related and structurally similar, but are significantly different in pixel values. The third column shows the average of these two patches. Each of the remaining column shows the reconstruction of the blending at a different layer. It is clear that pixel based and neural based blending give very different results: averaging the pixels often gives strong ghost artifacts. This can be reduced as we dive deeper into the network: through experiments we observed that the middle level layers such as \textit{relu3\_1} and \textit{relu4\_1} often give more meaningful blendings. Lower layers such as \textit{relu2\_1} behave similarly to pixels; reconstruction from layers beyond \textit{relu4\_1} tends to be too fuzzy to be used for synthesis. This also matches our previous observation of middle layers' privilege for patch matching -- because a representation that gives better discriminative performance is more robust to noise and enables better interpolation.

\begin{figure}[t]
	\centering
	\includegraphics[width=0.95\linewidth]{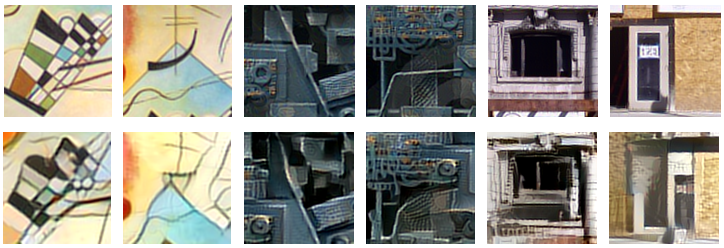}
	\caption{Effect of MRFs prior in neural based synthesis. We do not show the example patches here because they are almost visually identical to the synthesized patches cropped from our results (top). In contrast, patches cropped from~\cite{Gatys15}'s synthesis results have severe distortion and smear (bottom).}\label{fig:analysis_mrf}
\end{figure}

\subsection{Effect of the MRF Prior}
Despite the advantages in patch matching and blending, it is still possible for a dCNN to generate implausible results. For example, the matched patches from different layers might not always fire at the same place in the image space, indicated by the offsets between the patches found at layer 
\textit{relu3\_1} and \textit{relu4\_1} (Figure~\ref{fig:analysis_matching}). The blending may also produce artifacts, for example, the destructed face of the dog (\textit{relu4\_1}, figure~\ref{fig:analysis_blending}) and the ghosting eyes of the cat (\textit{relu3\_1}, figure~\ref{fig:analysis_blending}). Extreme cases can be found at~\cite{Mordvintsev15} which produces hallucinogenic images by simply stimulating neural activations without any constraint of the natural image statistics.

More natural results can be obtained through additional regularizer such as smothness (total variation) or the Gram matrix of pixel-wise filter responses~\cite{Mahendran15,Gatys15}.
Our approach adds an MRF regularizer to the middle / upper levels of the network.
%
To show the benefits of the MRFs prior, we compare the synthesis results with and without the constraint. We compare against the ``style constraint'' based on matching Gram matrices from~\cite{Gatys15}. Figure~\ref{fig:analysis_mrf} validates the intended improvements in local consistency. The first row shows image patches cropped from our results. They are visually consistent to the patches in the original style images. In contrast,~\cite{Gatys15} produces artifacts such as distortions and smears. The new MRF prior reduces flexibility in local adaptation in favor of reproducing meso-scale features more faithfully.

\section{Implementation Details}
This section describes implementation details of our algorithm~\footnote{We release code at: https://github.com/chuanli11/CNNMRF}. We use the pre-trained 19-layer VGG-Network from~\cite{Simonyan14c}. The synthesis $\bv{x}$ is initialized as random noise, and iteratively updated by minimizing Equation~\ref{eq:energy} using back-propagation. We use layer \textit{relu3\_1} and \textit{relu4\_1} for MRFs prior, and layer \textit{relu4\_2} for content constraint. 

For both layer \textit{relu3\_1} and \textit{relu4\_1} we use $3 \times 3$ patches. To achieve the best synthesis quality we set the stride to one so patches are very densely sampled. The patch matching (Equation~\ref{eq:ncorr}) is implemented as an additional convolutional layer for fast computation. In this case patches sampled from the style image are treated as the filters. The best matching of a query patch is the filter that gives the maximum response. We can pre-computed the magnitude of the filters ($||\Psi_i(\Phi(\bv{x}_{t}))||$ in Equation~\ref{eq:ncorr}) as they will not change during the synthesis. Unfortunately the magnitude of patches in the synthesized image ($||\Psi_i(\Phi(\bv{x}))||$) needs to be computed on the fly.

In practice we use a multi-resolution process: We built a image pyramid using the scaling factor of two, and stop when the longest dimension of the synthesized image is less than 64 pixels. The reference texture and reference content images are scaled accordingly. We perform 200 iterations for each resolution, and the output of the previous resolution is bi-linearly up-sampled as the initialization for the next resolution. We implement our algorithm under the Torch framework. Our algorithm take about three minutes to synthesis an image of size $384 \times 384$ with a Titan X GPU. 

To partially overcome the perspective and scale difference between the style and the content images, we sample patches from a number of copies of the style image with different rotations and scales: we use seven scales $\{0.85, 0.9, 0.95, 1, 1.05, 1.1, 1.15\}$, and for each scale we create five rotational copies $\{-\frac{\pi}{12}, -\frac{\pi}{24}, 0, \frac{\pi}{24}, \frac{\pi}{12}\}$. Since increasing the number of patches is computational expensive, in practice we only use the rotational copies for objects that can deform -- for example faces.

\begin{figure}[t]
	\centering
	\includegraphics[width=0.95\linewidth]{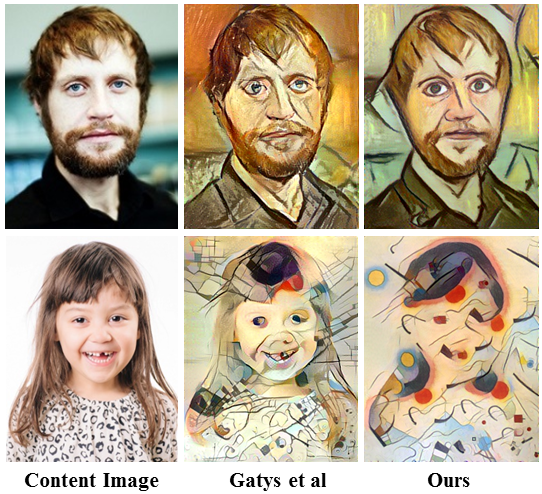}
	\caption{Comparison with Gatys et al.~\cite{Gatys15} for artistic synthesis. Content images credited to flickr users \emph{Christopher Michel} (top) and \emph{Peter Dahlgren} (bottom).}\label{fig:qlt_compare1}
\end{figure}

\section{Results}
This section discusses our results. Here we focus on style transfer and refer readers to our supplementary material for the results of un-guided synthesis. As discussed later in this section, our method has strong restriction with the input data, so a formal user study is less meaningful. For this reason, our evaluation is performed by demonstrating our method with a number of typical successful and failure cases, with comparisons and discussions related to the state of the art neural based image stylization method~\cite{Gatys15}. We refer readers to our supplementary report for more results.

Figure~\ref{fig:qlt_compare1} shows two examples of stylizing photos by artwork. The input photos (left) are stylized using Pablo Picasso's ``Self-portrait 1907'' and Wassily Kandinsky's ``Composition VIII'', respectively. In both cases, Gatys et al.'s~\cite{Gatys15} method (middle) achieves interesting results, preserving the content of the photos and the overall look and feel of the artworks. However, there are weaknesses on closer inspection: The first result (top row) contains many unnecessary details, and the eyes look unnatural. Their second result lost the characteristic shapes in the original painting and partially blends with the content exemplar. In contrast, our first result synthesized more plausible facial features. In the second result, our method also resembles the style better: notice the important facial features such as eyes and the mouth are synthesized as simple shapes, and the hair as regions of dark color.

\begin{figure*}[ht]
	\centering
	\includegraphics[width=0.95\linewidth]{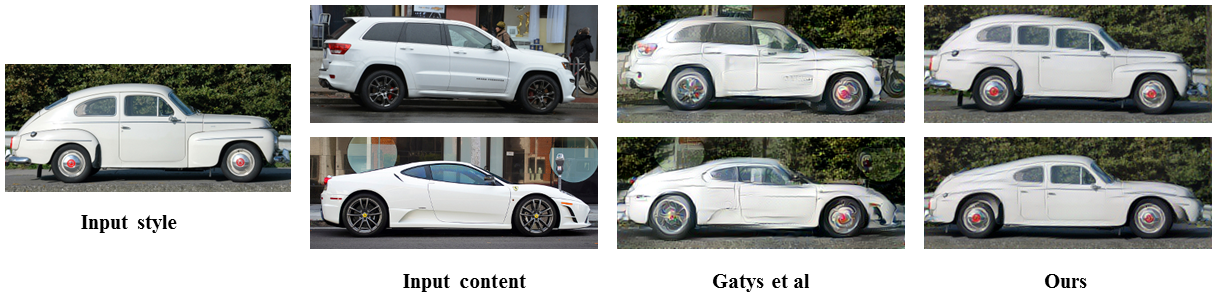}
	\caption{Comparison with Gatys et al.~\cite{Gatys15} for photo-realistic synthesis. Input images credited to flickr users \emph{Brett Levin}, \emph{Axion23} and \emph{Tim Dobbelaere}.}\label{fig:qlt_compare2}
\end{figure*}

Figure~\ref{fig:qlt_compare2} shows two examples of photorealsitic synthesis. We transfer the style of a vintage car to two different modern vehicles. Notice the lack of photo-realistic details and strong smears in~\cite{Gatys15}'s results. With the MRFs constraint (right) our results are closer to photorealistic.

\begin{figure*}[ht]
	\centering
	\includegraphics[width=0.95\linewidth]{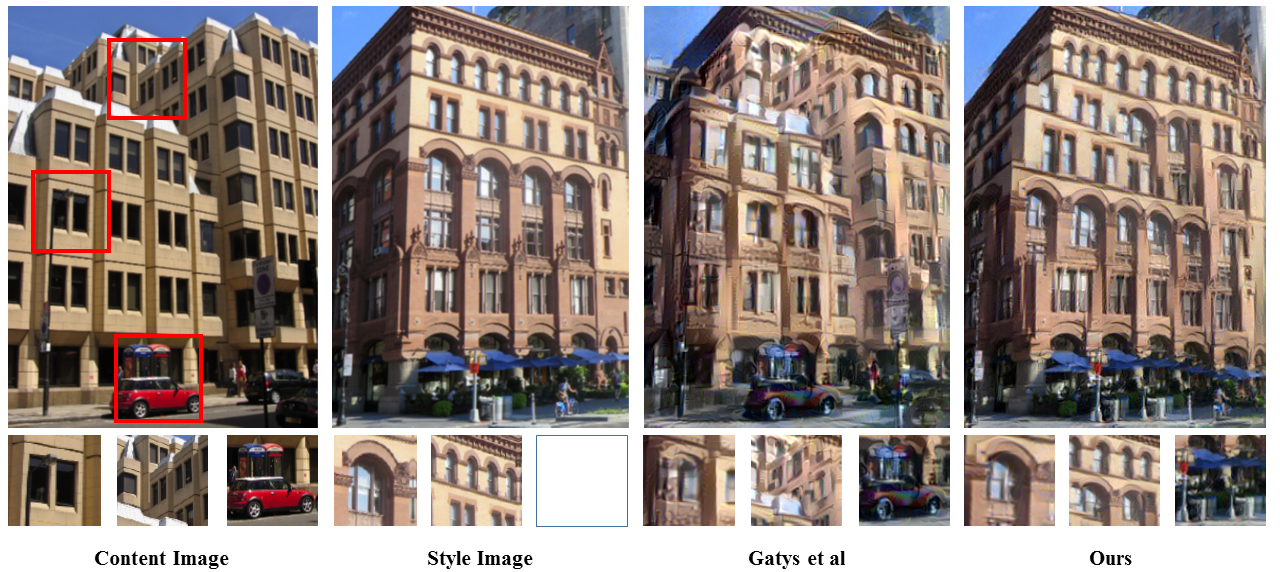}
	\caption{Detailed analysis of the difference between our results and Gatys et al.'s~\cite{Gatys15}'s results. Input images credited to flickr users \emph{Eden, Janine and Jim} and \emph{Garry Knight}.}\label{fig:qlt_compare3}
\end{figure*}

From an informal user study, we observed that \cite{Gatys15} usually keeps the content better, and our method produces more accurate styles. Figure~\ref{fig:qlt_compare3} gives detailed analysis between these two different characters. Here we show three patches from the content image (red boxes), and their closet matches from the style image and the synthesis images (using neural level matching). Notice our method produces more plausible results when a good match can be found between the content and the style images (the first patch), and performs less well when mis-matching happens (the second patch, where our synthesis deviates from the content image). For the third patch, there is no matching can be found for the car. In this case, our method replaces the car with texture synthesis, while \cite{Gatys15} keeps the car and renders it with artifacts. In general, our method creates more stylish images, but may contain artifacts when the MRFs do not fit the content. In contrast, the parametric method \cite{Gatys15} is more adaptable to the content, but at the cost of deviating from the style.

%

\subsection{Limitations}
Our method is an interesting extension for image based style transfer, especially for photorealistic styles. Nonetheless, it has many limitations. First and for most, it is more restricted to the input data: it only works if the content image can be re-assembled by the MRFs in the style image. For example, images of strong perspective or structure difference are not suitable for our method. 

\begin{figure}[t]
	\centering
	\includegraphics[width=0.95\linewidth]{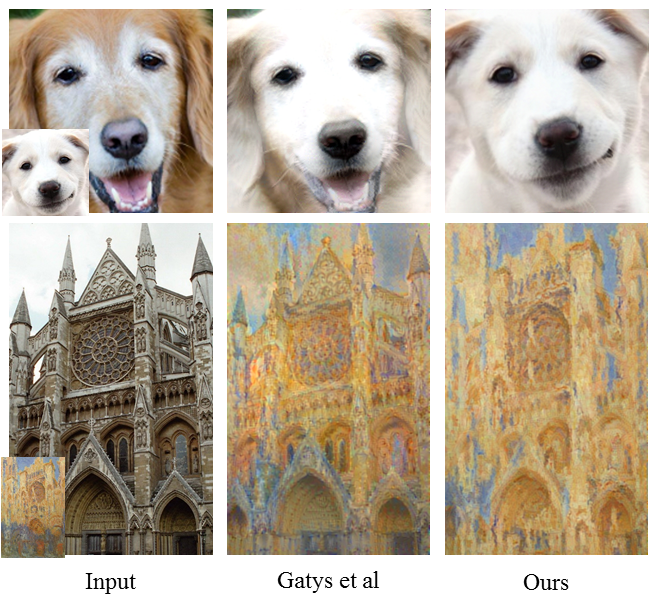}
	\caption{Cases where Gatys et al ~\cite{Gatys15} works better.}\label{fig:limit}
\end{figure}

Figure~\ref{fig:limit} shows two further example cases where~\cite{Gatys15} works better: The first example aim to transfer the style of a white dog according to the content of a yellow dog. Notice how the dogs' facial expression differ in these images. In this case our result (right) failed to reproduce the open mouth in the content image. The second example shows a artistic style that can be more easily transferred by the parametric method: Notice how the textures adapt to the content more naturally in~\cite{Gatys15}'s method. In contrast, our method tries to ``reshuffle'' building blocks in the style image and lost important features in the content image. In general, our method works better for subjects that allows structural deformation, such as faces and cars. For subjects that have strict symmetry properties such as architectures, it is often that our method will generate structural artifacts. In this case, structural statistics such as~\cite{He12} may be used for regularizing the synthesized image.

Although our method achieved improvement for photorealistic synthesis, it is still not as sharp as the original photos. This is due to the loss of non-discriminative image details during the training of the network. This opens an interesting future work that how the dCNN can be retrained, or incorporated with stitching based texture synthesis such as~\cite{Kwatra03} for achieving pixel-level photorealism.

\section{Conclusions}
The key insight of this paper is that we can combine the discriminative power of a deep neural network with classical MRFs based texture synthesis. We developed a simple method that is able to produce encouraging new results for style transfer between images. We analyzed our results with a number of typical successful and failure cases, and discussed its pros and cons compared to the state of the art neural based method for transferring image styles. Importantly, our results often preserve better mesostructures in the synthesized image. For this first time, this permits transferring photo-realistic styles with some plausibility. The stricter control of the mesostructure is also the biggest limitation at this point: The MRF prior only offers advantages when style and content images consists of similarly shaped elements without strong changes in perspective, size, or shape, as covered by the invariance of the high-level neural encoding. Otherwise, artifacts might occur. For pure artistic styles, the increased rigidity can then be a disadvantage.

Our work is only one step in the direction of leveraging deep convolutional neural networks for improving image synthesis. It opens many interesting questions and future work such as how to resolve the incompatibility between the structure guidance and the MRFs~\cite{He12}; how to more efficiently study and transfer the middle level style across a big dataset; and how to generate pixel-level photorealistic images by incorporating with stitching based texture synthesis~\cite{Kwatra03}, or with generative training of the network.

\section*{Acknowledgments}
This work has been partially supported by the Intel Visual Computing Institute and by the International Max Planck Research School for Computer Science. We thank Hao Su and Daniel Franzen for inspiring discussions.

{\small
\bibliographystyle{ieee}
\bibliography{paper}
}

\end{document}